\documentclass[10pt,twocolumn,letterpaper]{article}

\usepackage{cvpr}              

\usepackage{graphicx}
\usepackage{amsmath}
\usepackage{amssymb}
\usepackage{booktabs}

\usepackage{bm}
\usepackage{multirow}
\usepackage{subcaption}
\usepackage{algorithm}
\usepackage{algpseudocode}
\usepackage{float}
\usepackage{capt-of}
\usepackage[numbers,sort,compress]{natbib}
\usepackage[dvipsnames]{xcolor}
\usepackage{lipsum}
\usepackage{bbm}

\DeclareMathOperator*{\argmin}{arg\,min}

\usepackage[pagebackref,breaklinks,colorlinks]{hyperref}

\usepackage[capitalize]{cleveref}
\crefname{section}{Sec.}{Secs.}
\Crefname{section}{Section}{Sections}
\Crefname{table}{Table}{Tables}
\crefname{table}{Tab.}{Tabs.}

\newfloat{algorithm}{t}{lop}

\newcommand*{\ours}{MotionDiffuser}
\newcommand*{\var}[1]{\text{Var}(#1)}
\newcommand*{\cskip}{c_{\text{skip}}}
\newcommand*{\cin}{c_{\text{in}}}
\newcommand*{\cout}{c_{\text{out}}}
\newcommand*{\cnoise}{c_{\text{noise}}}

\definecolor{amber500}{rgb}{1.0, 0.804, 0.333}
\definecolor{coral500}{rgb}{0.957, 0.686, 0.569}
\definecolor{purple500}{rgb}{0.568, 0.376, 0.796}

\def\cvprPaperID{11647} 
\def\confName{CVPR}
\def\confYear{2023}

\makeatletter
\newcommand\notsotiny{\@setfontsize\notsotiny\@vipt\@viipt}
\makeatother

\begin{document}

\title{\ours{}: Controllable Multi-Agent Motion Prediction using Diffusion}

\author{Chiyu ``Max" Jiang$^{*}$ \qquad Andre Cornman$^{*}$ \qquad Cheolho Park \\
\qquad Ben Sapp \qquad Yin Zhou \qquad Dragomir Anguelov \\
{\tt\small {$*$ equal contribution}} \\
Waymo LLC \\
}

\setlength{\abovedisplayskip}{0.2\abovedisplayskip}
\setlength{\belowdisplayskip}{0.2\belowdisplayskip}
\setlength{\abovecaptionskip}{0.2\abovecaptionskip}
\setlength{\belowcaptionskip}{0.2\belowcaptionskip}

\twocolumn[{%
\renewcommand\twocolumn[1][]{#1}%
\maketitle
\vspace{-3em}
\begin{center}
    \centering
    \centering
    \includegraphics[width=.9\linewidth]{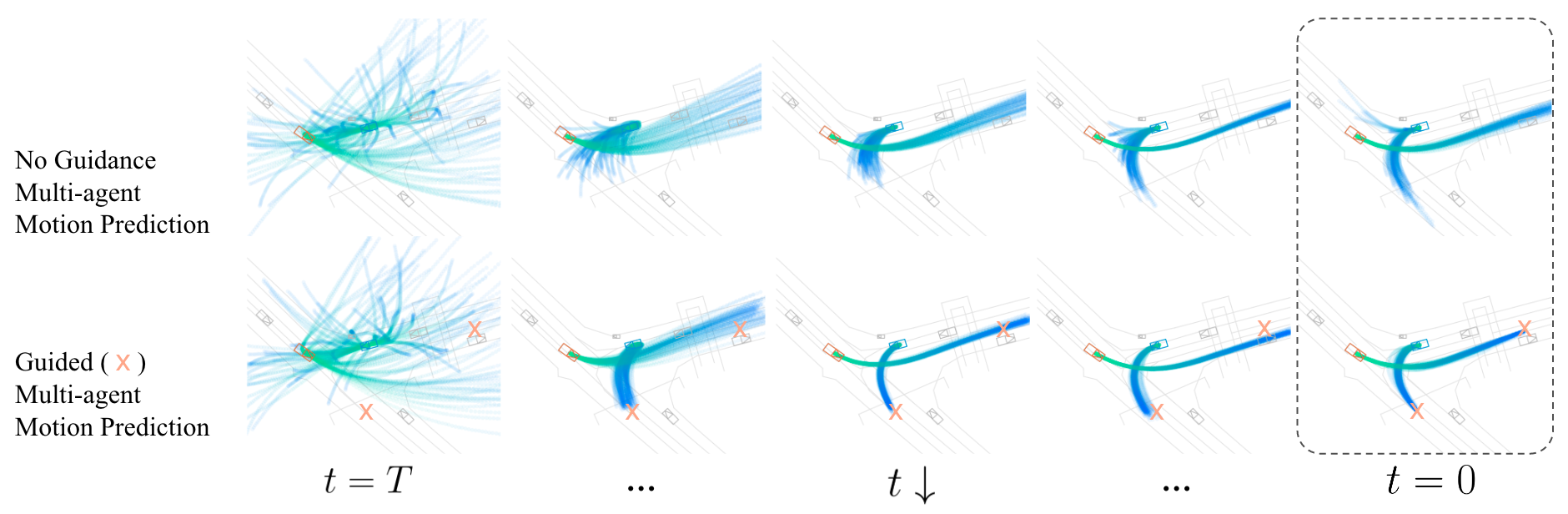}
    \captionof{figure}{\ours{} is a learned representation for the distribution of multi-agent trajectories based on diffusion models. During inference, samples from the predicted joint future distribution are first drawn i.i.d. from a random normal distribution (leftmost column), and gradually denoised using a learned denoiser into the final predictions (rightmost column). Diffusion allows us to learn a diverse, multimodal distribution over joint outputs (top right). Furthermore, guidance in the form of a differentiable cost function can be applied at inference time to obtain results satisfying additional priors and constraints (bottom right).}
    \label{fig:teaser}
\end{center}%

}]

\begin{abstract}
\vspace{-1em}

    We present \ours{}, a diffusion based representation for the joint distribution of future trajectories over multiple agents. Such representation has several key advantages:  first, our model learns a highly multimodal distribution that captures diverse future outcomes. Second, the simple predictor design requires only a single L2 loss training objective, and does not depend on trajectory anchors. Third, our model is capable of learning the joint distribution for the motion of multiple agents in a permutation-invariant manner. Furthermore, we utilize a compressed trajectory representation via PCA, which improves model performance and allows for efficient computation of the exact sample log probability. Subsequently, we propose a general  constrained sampling framework that enables controlled trajectory sampling based on differentiable cost functions. This strategy enables a host of applications such as enforcing rules and physical priors, or creating tailored simulation scenarios. \ours{} can be combined with existing backbone architectures to achieve top motion forecasting results. We obtain state-of-the-art results for multi-agent motion prediction on the Waymo Open Motion Dataset.

\end{abstract}

\section{Introduction}
\label{sec:intro}
Motion prediction is a central yet challenging problem for autonomous vehicles to safely navigate under uncertainties. Motion prediction, in the autonomous driving setting, refers to the prediction of the future trajectories of modeled agents, conditioned on the histories of the modeled agents, context agents, road graph and traffic light signals.

\begin{figure*}
    \centering
    \includegraphics[width=\linewidth]{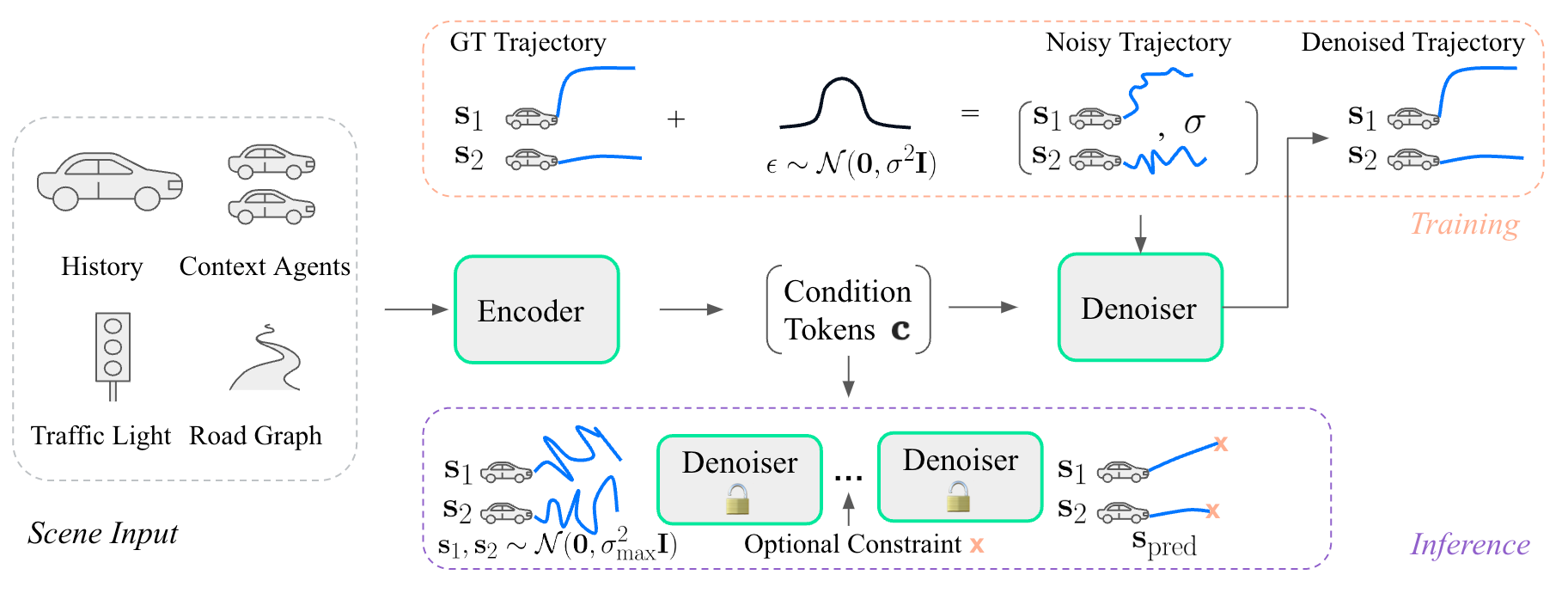}
    \caption{Overview for multi-agent motion prediction using diffusion models. The input scene containing agent history, traffic lights and road graphs is encoded via a transformer encoder into a set of condition tokens $\bm{C}$. During \textcolor{coral500}{training}, a random set of noises are sampled i.i.d. from a normal distribution and added to the ground truth (GT) trajectory. The denoiser, while attending to the condition tokens, predicts the denoised trajectories corresponding to each agent. The entire model can be trained end-to-end using a simple L2 loss between the predicted denoised trajectory and the GT trajectory. During \textcolor{purple500}{inference}, a population of trajectories for each agent can first be sampled from pure noise at the highest noise level $\sigma_{\text{max}}$, and iteratively denoised by the denoiser to produce a plausible distribution of future trajectories. An optional constraint in the form of an arbitrary differentiable loss function can be injected in the denoising process to enforce constraints.}
    \label{fig:overview}
\end{figure*}

Several key challenges arise in the motion prediction problem. First, motion prediction is probabilistic and multi-modal in nature where it is important to faithfully predict an unbiased distribution of possible futures. Second, motion prediction requires jointly reasoning about the future distribution for a set of agents that may interact with each other in each such futures. Naively predicting and sampling from the marginal distribution of trajectories for each agent independently leads to unrealistic and often conflicting outcomes. Last but not least, while it is challenging to constrain or bias the predictions of conventional regression-based trajectory models, guided sampling of the trajectories is often required. For example, it may be useful to enforce rules or physical priors for creating tailored simulation scenarios. This requires the ability to enforce constraints over the future time steps, or enforce a specified behavior for one or more agents among a set of agents.

In light of these challenges, we present \ours{}, a denoising diffusion model-based representation for the joint distribution of future trajectories for a set of agents (see Fig. \ref{fig:overview}). \ours{} leverages a conditional denoising diffusion model. Denoising diffusion models \cite{ho2020denoising,nichol2021improved,song2020score,song2020denoising,karras2022elucidating} (henceforth, diffusion models) are a class of generative models that learns a denoising function based on noisy data and samples from a learned data distribution via iteratively refining a noisy sample starting from pure Gaussian noise (see Fig. \ref{fig:teaser}). Diffusion models have recently gained immense popularity due to their simplicity, strong capacity to represent complex, high dimensional and multimodal distributions, ability to solve inverse problems \cite{song2020score,choi2021ilvr,chung2022come,kawar2022denoising}, and effectiveness across multiple problem domains, including image generation \cite{ramesh2022hierarchical,saharia2022photorealistic,rombach2022high}, video generation \cite{ho2022video,yang2022diffusion,ho2022imagen} and 3D shape generation \cite{poole2022dreamfusion}.

Building on top of conditional diffusion models as a basis for trajectory generation, we propose several unique design improvements for the multi-agent motion prediction problem. First, we propose a cross-attention-based permutation-invariant denoiser architecture for learning the motion distribution for a set of agents regardless of their ordering. Second, we propose a general and flexible framework for performing controlled and guided trajectory sampling based on arbitrary differentiable cost functions of the trajectories, which enables several interesting applications such as rules and controls on the trajectories, trajectory in-painting and creating tailored simulation scenarios. Finally, we propose several enhancements to the representation, including PCA-based latent trajectory diffusion and improved trajectory sample clustering to further boost the performance of our model.

In summary, the main contributions of this work are:
\begin{itemize}
    \item A novel permutation-invariant, multi-agent joint motion distribution representation using conditional diffusion models.
    \item A general and flexible framework for performing controlled and guided trajectory sampling based on arbitrary differentiable cost functions of the trajectories with a range of novel applications.
    \item Several significant enhancements to the representation, including PCA-based latent trajectory diffusion formulation and improved trajectory sample clustering algorithm to further boost the model performance.
\end{itemize}

\section{Related Work}
\label{sec:related_work}
\paragraph{Denoising diffusion models} Denoising diffusion models \cite{ho2020denoising,nichol2021improved}, methodologically highly related to the class of score-based generative models \cite{song2020score,song2020denoising,karras2022elucidating}, have recently emerged as a powerful class of generative models that demonstrate high sample quality across a wide range of application domains, including image generation \cite{ramesh2022hierarchical,saharia2022photorealistic,rombach2022high}, video generation \cite{ho2022video,yang2022diffusion,ho2022imagen} and 3D shape generation \cite{poole2022dreamfusion}. We are among the first to use diffusion models for predicting the joint motion of agents.
\paragraph{Constrained sampling} Diffusion models have been shown to be effective at solving inverse problems such as image in-painting, colorization and sparse-view computed tomography by using a controllable sampling process \cite{song2020score, choi2021ilvr, chung2022come, kawar2022denoising, chung2022improving, kadkhodaie2021stochastic, song2020denoising}. Concurrent work \cite{zhong2022guided} explores diffusion modeling for controllable traffic generation, which we compare to in Sec. \ref{ssec:constrained_traj_sample}. In diffusion models, the generation process can be conditioned on information not available during training. The inverse problem can be posed as sampling from the posterior $p(\bm{x};\bm{y})$ based on a learned unconditional distribution $p(\bm{x})$, where $\bm{y}$ is an observation of the event $\bm{x}$. We defer further technical details to Sec. \ref{ssec:constrained_traj_sample}.

\paragraph{Motion prediction}
There are two main categories of approaches for motion prediction: supervised learning and generative learning. Supervised learning trains a model with logged trajectories with supervised losses such as L2 loss. One of the challenges is to model inherent multi-modal behavior of the agents. For this, MultiPath \cite{sapp2020multipath} uses static anchors, and MultiPath++ \cite{varadarajan2021multipath++}, Wayformer \cite{nayakanti2022wayformer}, SceneTransformer \cite{ngiam21scene_transformer} use learned anchors, and DenseTNT \cite{gu21dense_tnt} uses goal-based predictions. Home \cite{gilles2021home} and GoHome \cite{gilles2022gohome} predict future occupancy heatmaps, and then decode trajectories from the samples. MP3 \cite{casas2021mp3} and NMP \cite{zeng2019end} learn the cost function evaluator of trajectories, and then the output trajectories are heuristically enumerated. Many of these approaches use ensembles for further diversified predictions. The next section covers generative approaches.


\paragraph{Generative models for motion prediction} Various recent works have modeled the motion prediction task as a conditional probability inference problem of the form $p(\bm{s};\bm{c})$ using generative models, where $\bm{s}$ denote the future trajectories of one or more agents, and $\bm{c}$ denote the context or observation.
HP-GAN \cite{barsoum2018hp} learns a probability density function (PDF) of future human poses conditioned on previous poses using an improved Wasserstein Generative Adversarial Network (GAN). Conditional Variational Auto-Encoders (C-VAEs) \cite{ivanovic2020multimodal,oh2022cvae,gomez2020real}, Normalizing Flows \cite{scholler2021flomo, fadel2021contextual, mao2021generating, ma2020diverse} have also been shown to be effective at learning this conditional PDF of future trajectories for motion prediction. Very recent works have started looking into diffusion models as an alternative to modeling the conditional distributions of future sequences such as human motion pose sequences \cite{saadatnejad2022generic,zhang2022motiondiffuse} and planning \cite{janner2022diffuser}. In a more relevant work, \cite{gu2022stochastic} the authors utilize diffusion models to model the uncertainties of pedestrian motion. As far as we are aware, we are the first to utilize diffusion models to model the multi-agent joint motion distribution.

\paragraph{Multi-agent motion prediction}
While much of the motion prediction literature has worked on predicting motions of individual agents independently, there has been some work to model the motion of multiple agents jointly. SceneTransformer \cite{ngiam21scene_transformer} outputs a fixed set of joint motion predictions for all the agents in the scene. M2I \cite{sun2022m2i}, WIMP \cite{khandelwal2020if}, PIP \cite{song2020pip}, and CBP \cite{tolstaya2021cbp} propose a conditional model where the motions of the other agents are predicted by given motions of the controlled agents.

There is a set of literature using probabilistic graphical models. DSDNet \cite{zeng2020dsdnet} and MFP \cite{tang_multifuture} use fully connected graphs. JFP \cite{wenjie2022} supports static graphs such as fully connected graphs and autonomous vehicle centered graphs, and dynamic graphs where the edges are constructed between the interacting agents. RAIN \cite{li2021rain} learns the dynamic graph of the interaction through separate RL training.

\vspace{-.5em}
\section{Method}
\label{sec:method}
\subsection{Diffusion Model Preliminaries}
\paragraph{Preliminaries} Diffusion models \cite{karras2022elucidating} provide a learned parameterization of the probability distribution $p_{\bm{\theta}}(\bm{x})$ through learnable parameters $\bm{\theta}$. Denote this probability density function, convolved with a Gaussian kernel of standard deviation $\sigma$ to be $p_{\bm{\theta}}(\bm{x}, \sigma)$. Instead of directly learning a normalized probability density function $p_{\bm{\theta}}(\bm{x})$ where the normalization constant is generally intractable \cite{hyvarinen2005estimation}, diffusion models learn the score function of the distribution: $\nabla_{\bm{x}}\log p_{\bm{\theta}}(\bm{x}; \sigma)$ at a range of noise levels $\sigma$.

Given the score function $\nabla_{\bm{x}}\log p_{\bm{\theta}}(\bm{x}; \sigma)$, one can sample from the distribution by denoising a noise sample. Samples can be drawn from the underlying distribution $\bm{x}_0 \sim p_{\bm{\theta}}(\bm{x})$ via the following dynamics:
\begin{align}
    \bm{x}_0 = \bm{x}(T)+\int_{T}^{0} -\dot{\sigma}(t) \sigma(t) \nabla_{\bm{x}}\log p_{\bm{\theta}}(\bm{x}(t); \sigma(t)) dt \nonumber \\
    \text{where} \quad \bm{x}(T) \sim \mathcal{N}(\bm{0}, \sigma_{
    \text{max}}^2 \bm{I})
    \label{eqn:ode}
\end{align}
where variance $\sigma(t)$ is a monotonic, deterministic function of an auxiliary parameter of time $t$. Following \cite{karras2022elucidating}, we use the linear noise schedule $\sigma(t)=t$. The initial noise sample is sampled i.i.d. from a unit Gaussian scaled to the highest standard deviation $\sigma(T) = \sigma_{\text{max}}$.

The diffusion model can be trained to approximate a data distribution $p_{\chi}(\bm{x})$, where $\chi = \{\bm{x}_1, \bm{x}_2, \cdots, \bm{x}_{Nd}\}$ denote the set of training data. The empirical distribution of the data can be viewed as a sum of delta functions around each data point: $p_{\chi}(\bm{x}) = \frac{1}{n}\sum_{i=0}^{N_d} \delta(\bm{x}-\bm{x}_i)$. Denote the denoiser as $D(\bm{x};\sigma)$ which is a function that recovers the unnoised sample corresponding to the noised sample $\bm{x}$. The denoiser is related to the score function via:
\begin{align}
    \nabla_{\bm{x}}\log p(\bm{x};\sigma)=(D(\bm{x};\sigma)-\bm{x})/\sigma^2
    \label{eqn:score_vs_denoiser}
\end{align}
The denoiser can be learned by minimizing the expected $L_2$ denoising error for a perturbed sample $\bm{x}$ at any noise level $\sigma$ sampled from the noise distribution $q(\sigma)$:
\begin{align}
    \argmin_{\theta} ~ \mathbb{E}_{\bm{x}\sim p_{\chi}}\mathbb{E}_{\sigma\sim q(\sigma)}\mathbb{E}_{\bm{\epsilon}\sim\mathcal{N}(\bm{0},\sigma^2\bm{I})}\mathbb||D_{\bm{\theta}}(\bm{x}+\bm{\epsilon};\sigma)-\bm{x}||_2^2
\end{align}

\begin{figure}
    \centering
    \includegraphics[width=\linewidth]{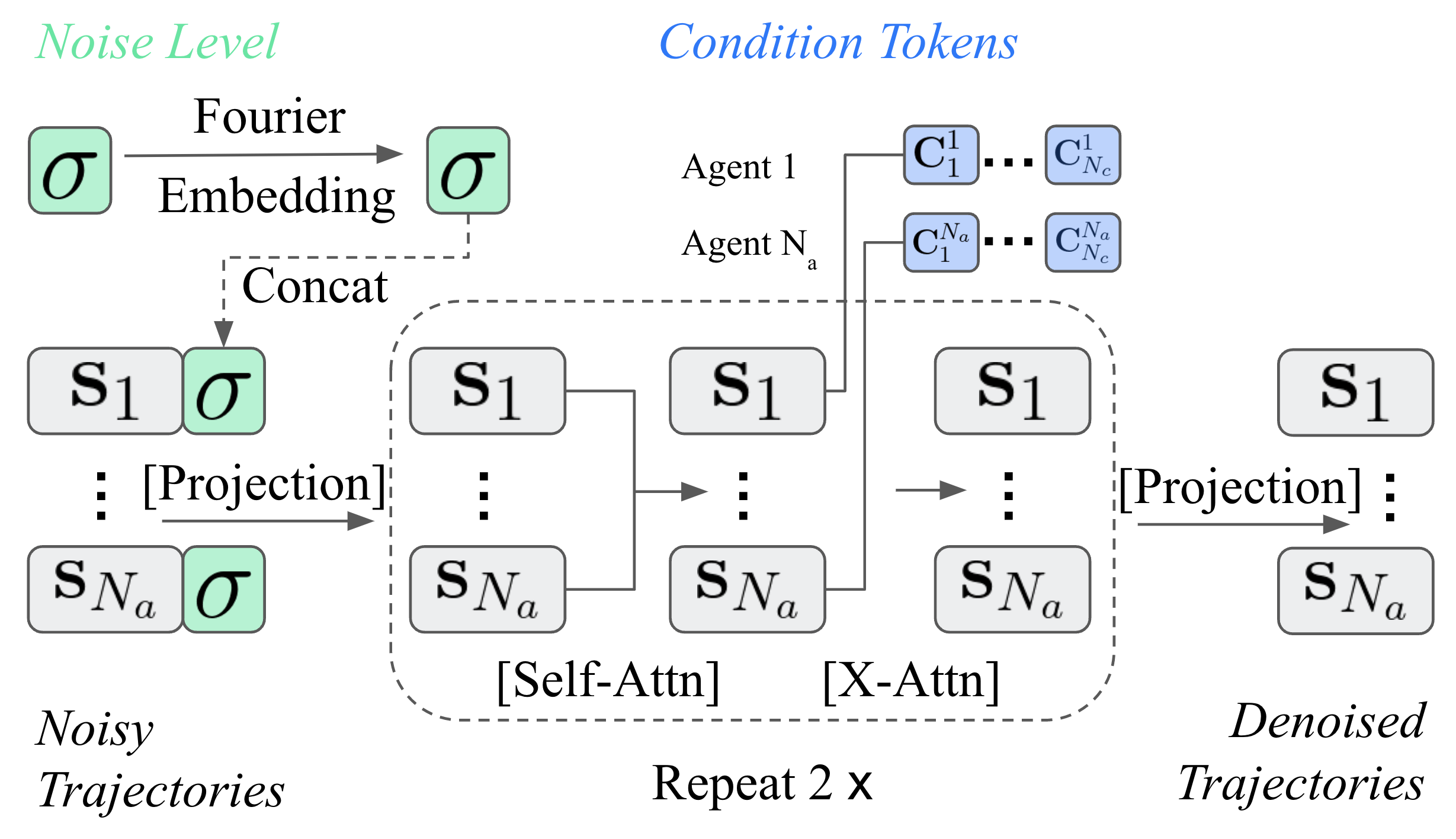}
    \caption{Network architecture for set denoiser $D_{\bm{\theta}}(\bm{S};\bm{C},\sigma)$. The noisy trajectories corresponding to agents $\bm{s}_1\cdots\bm{s}_{N_a}$ are first concatenated with a random-fourier encoded noise level $\sigma$, before going through repeated blocks of self-attention among the set of trajectories and cross-attention with respect to the condition tokens $\bm{c}_1\cdots\bm{c}_{N_c}$. The self-attention allows the diffusion model to learn a joint distribution across the agents and cross-attention allows the model to learn a more accurate scene-conditional distribution. Note that each agent cross-attends to its own condition tokens from the agent-centric scene encoding (not shown for simplicity). The [learnable components] are marked with brackets.} 
    \label{fig:denoiser_architecture}
    \vspace{-1em}
\end{figure}

\paragraph{Conditional diffusion models} In this work, we are interested in the conditional setting of learning $p_{\bm{\theta}}(\bm{x};\bm{c})$, where $\bm{x}$ denote the future trajectories of a set of agents and $\bm{c}$ is the scene context. A simple modification is to augment both the denoiser $D(\bm{x};\bm{c},\sigma)$ and the score function $\nabla_{\bm{x}}\log p(\bm{x};\bm{c},\sigma)$ by the condition $\bm{c}$. Given a dataset $\chi_c$ augmented by conditions: $\chi_c=\{(\bm{x}_1, \bm{c}_1), \cdots, (\bm{x}_{N_d}, \bm{c}_N)\}$, the conditional denoiser can be learned by a conditional denoising score matching objective to minimize the following:
\begin{align}
    \mathbb{E}_{\bm{x},\bm{c}\sim \chi_c}\mathbb{E}_{\sigma\sim q(\sigma)}\mathbb{E}_{\bm{\epsilon}\sim\mathcal{N}(\bm{0},\sigma^2\bm{I})}\mathbb||D_{\bm{\theta}}(\bm{x}+\bm{\epsilon};\bm{c},\sigma)-\bm{x}||_2^2
    \label{eqn:denoiser_loss}
\end{align}
which leads to the learned conditional score function:
\begin{align}
    \nabla_{\bm{x}}\log p_{\bm{\theta}}(\bm{x};\bm{c},\sigma)=(D_{\bm{\theta}}(\bm{x};\bm{c},\sigma)-\bm{x})/\sigma^2
    \label{eqn:cond_score_fn}
\end{align}

\vspace{-1em}
\paragraph{Preconditioning and training} Directly training the model with the denoising score matching objective (Eqn. \ref{eqn:denoiser_loss}) has various drawbacks. First, the input to the denoiser has non-unit variance: $\var{\bm{x}+\bm{\epsilon}}=\var{\bm{x}}+\var{\bm{\epsilon}}=\sigma_{\text{data}}^2+\sigma^2,~\sigma\in[0,\sigma_{\text{max}}]$. Second, at small noise levels of $\sigma$, it is much easier for the model to predict the residual noise than predicting the clean signal. Following \cite{karras2022elucidating}, we adopt a preconditioned form of the denoiser:
\begin{align}
    D_{\bm{\theta}}(\bm{x};\bm{c},\sigma)=\cskip(\sigma)\bm{x}+\cout(\sigma)F_{\bm{\theta}}(\cin(\sigma)\bm{x};\bm{c},\cnoise(\sigma))
\end{align}
$F_{\bm{\theta}}$ is the neural network to train, $\cskip, \cin, \cout, \cnoise$ respectively scale the skip connection to the noisy $\bm{x}$, input to the network, output from the network, and noise input $\sigma$ to the network. We do not additionally scale $\bm{c}$ since it is the output of an encoder network, assumed to have modulated scales.

\paragraph{Sampling} We follow the ODE dynamics in Eqn. \ref{eqn:ode} when sampling the predictions. We utilize Huen's 2nd order method for solving the corresponding ODE using the default parameters and 32 sampling steps.

\subsection{Diffusion Model for Multi-Agent Trajectories}
One of the main contributions of this work is to propose a framework for modeling the joint distribution of multi-agent trajectories using diffusion models. Denote the future trajectory of agent $i$ as $\bm{s}_i \in \mathbb{R}^{N_t\times N_f}$ where $N_t$ is the number of future time steps and $N_f$ is the number of features per time steps, such as longitudinal and lateral positions, heading directions etc. Denote $\bm{c}_i\in\mathbb{R}^{\cdots}$ as the learned ego-centric context encoding of the scene, including the road graph, traffic lights, histories of modeled and context agents, as well as interactions within these scene elements, centered around agent $i$. For generality $\bm{c}$ could be of arbitrary dimensions, either as a single condition vector, or as a set of context tokens. Denote the set of agent futures trajectories as $\bm{S}\in\mathbb{R}^{N_a\times N_t\times N_f}$, the set of ego-centric context encodings as $\bm{C} \in \mathbb{R}^{N_a\times\cdots}$, where $|\mathcal{S}| = |\mathcal{C}| = N_a$ is the number of modeled agents. We append each agent’s position and heading (relative to the ego vehicle) to its corresponding context vectors.Denote the $j$-th permutation of agents in the two sets to be $\mathcal{S}^j, \mathcal{C}^j$, sharing consistent ordering of the agents. We seek to model the set probability distribution of agent trajectories using diffusion models: $p(\mathcal{S}^j ; \mathcal{C}^j)$. Since the agent ordering in the scene is arbitrary, learning a permutation invariant set probability distribution is essential, i.e.,
\begin{align}
    p(\bm{S};\bm{C}) &= p(\bm{S}^j;\bm{C}^j), \forall j \in [1, N_a!]\label{eqn:set_prob}
\end{align}

To learn a permutation-invariant set probability distribution, we seek to learn a permutation-\textit{equivariant} denoiser, i.e., when the order of the agents in the denoiser permutes, the denoiser output follows the same permutation:
\begin{align}
    D(\bm{S}^j; \bm{C}^j, \sigma) = D^j(\bm{S}; \bm{C}, \sigma), \forall j\in [1, N_a!]
\end{align}
Another major consideration for the denoiser architecture is the ability to effectively attend to the condition tensor $\bm{c}$ and noise level $\sigma$. Both of these motivations prompt us to utilize the transformer as the main denoiser architecture. We utilize the scene encoder architecture from the state-of-the-art Wayformer \cite{nayakanti2022wayformer} model to encode scene elements such as road graph, agent histories and traffic light states into a set of latent embeddings. The denoiser takes as input the GT trajectory corresponding to each agent, perturbed with a random noise level $\sigma \sim q(\sigma)$, and the noise level $\sigma$. During the denoising process, the noisy input undergoes repeated blocks of self-attention between the agents and cross attention to the set of context tokens per agent, and finally the results are projected to the same feature dimensionality as the inputs. Since we do not apply positional encoding along the agent dimension, transformers naturally preserve the equivariance among the tokens (agents), leading to the permutation-equivarnance of the denoiser model. See Fig. \ref{fig:denoiser_architecture} for a more detailed design of the transformer-based denoiser architecture.

\begin{figure}
    \centering
    \includegraphics[width=\linewidth]{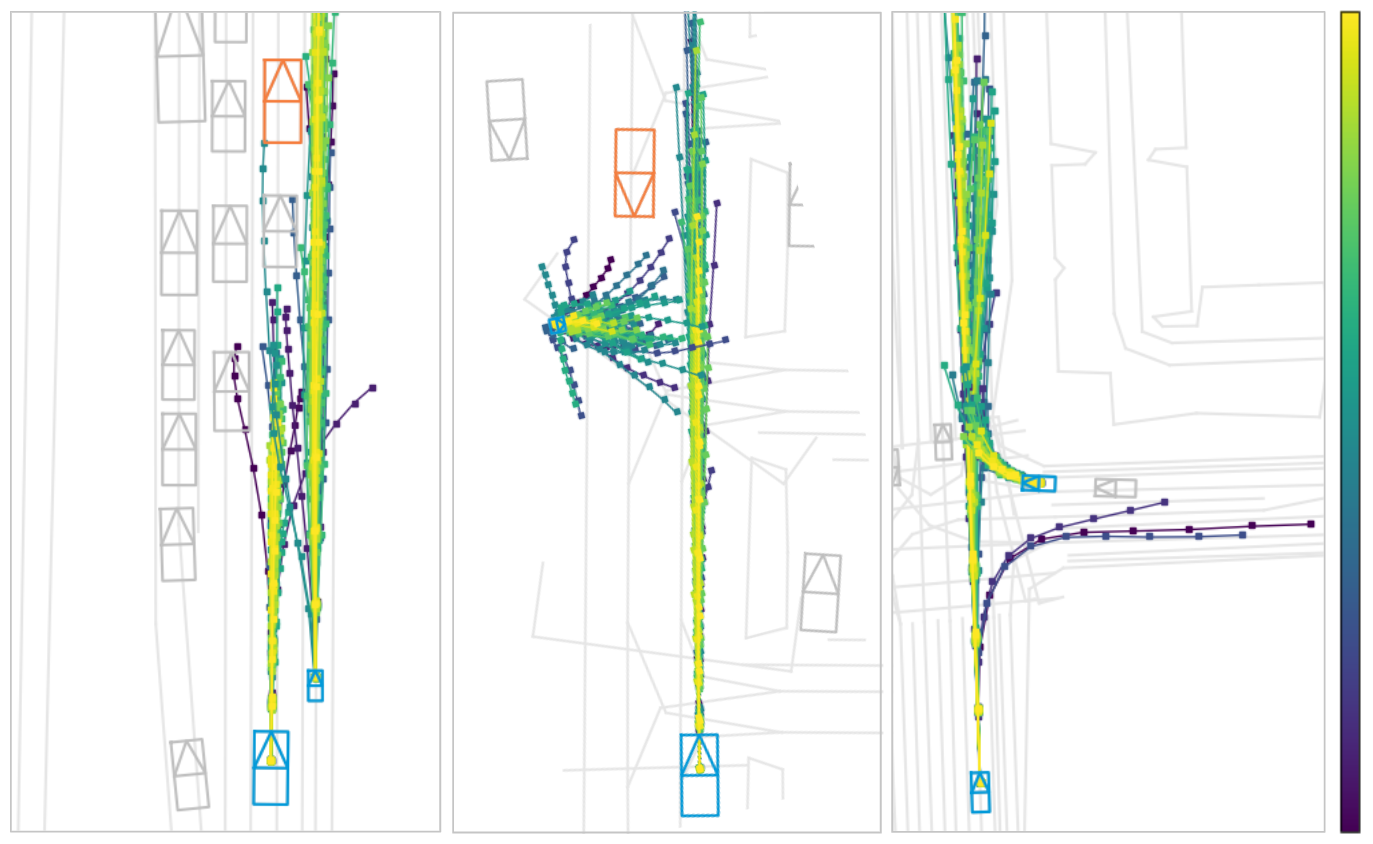}
    \caption{Inferred exact log probability of 64 sampled trajectories per agent. Higher probability samples are plotted with lighter colors. The orange agent represents the AV (autonomous vehicle).}
    \label{fig:traj_exact_logprob}
\end{figure}

\subsection{Exact Log Probability Inference}

With our model, we can infer the exact log probability of the generated samples with the following method. First, the change of log density over time follows a second differential equation, called the instantaneous change of variables formula \cite{chen2018neural},

\begin{align}
    \frac{\log p(\bm{x}(t))}{\partial t} = -Tr \left( \frac{\partial f}{\partial \bm{x}(t)} \right) \nonumber \\
    \text{where} f = \partial \bm{x} / \partial t 
    \label{eqn:instant_change_of_var}
\end{align}


In the diffusion model, the flow function, $f$ follows,

\begin{align}
    f(\bm{x}(t), t) = \frac{\partial \bm{x}(t)}{\partial t}= -\dot{\sigma}(t) \sigma(t) \nabla_{\bm{x}}\log p(\bm{x}(t); \sigma(t)) 
    \label{eqn:flow_function}
\end{align}

The log probability of the sample can be calculated by integrating over time as below.

\begin{align}
    \log p(\bm{x}(0)) = \log p(\bm{x}(T)) - \int^{0}_{T} Tr \left( \frac{\partial f}{\partial \bm{x}(t)} \right) dt 
    \label{eqn:log_prob}
\end{align}

The computation of the trace of the Jacobian takes $O(n^2)$ where $n$ is the dimensionality of $\bm{x}$. When we use PCA as in Sec. \ref{ssec:representation_enhancements}, $n$ will be much smaller than the dimensionality of the original data. We can also use Hutchinson's trace estimator as in FFJORD \cite{grathwohl2018ffjord} which takes $O(n)$.

The log probability can be used for filtering higher probability predictions. In Fig. \ref{fig:traj_exact_logprob}, for example, higher probability samples plotted with lighter colors are more likely.

\subsection{Constraining Trajectory Samples}\label{ssec:constrained_traj_sample}
Constrained trajectory sampling has a range of applications. One situation where controllability of the sampled trajectories would be required is to inject physical rules and constraints. For example, agent trajectories should avoid collision with static objects and other road users. Another application is to perform trajectory in-painting: to solve the inverse problem of completing the trajectory prediction given one or more control points. This is a useful tool in creating custom traffic scenarios for autonomous vehicle development and simulation.

More formally, we seek the solution to sampling from the joint conditional distribution $p(\bm{S};\bm{C})\cdot q(\bm{S};\bm{C})$, where $p(\bm{S};\bm{C})$ is the learned future distribution for trajectories and $q(\bm{S};\bm{C})$ a secondary distribution representing the constraint manifold for $\bm{S}$. The score of this joint distribution is $\nabla_{\bm{S}}\log \Big(p(\bm{S};\bm{C})\cdot q(\bm{S};\bm{C})\Big) = \nabla_{\bm{S}}\log p(\bm{S};\bm{C})+\nabla_{\bm{S}}\log q(\bm{S};\bm{C})$. In order to sample this joint distribution, we need the joint score function at all noise levels $\sigma$:
\begin{align}
    \nabla_{\bm{S}}\log p(\bm{S};\bm{C},\sigma)+\nabla_{\bm{S}}\log q(\bm{S};\bm{C},\sigma)
\end{align}
The first term directly corresponds to the conditional score function in Eqn. \ref{eqn:cond_score_fn}. The second term accounts for gradient guidance based on the constraint, which resembles classifier-based guidance \cite{ho2022classifier} in class-conditional image generation tasks, where a specialty neural network is trained to estimate this guidance term under a range of noise levels. We refer to this as the constraint gradient score. However, since our goal is to approximate the constraint gradient score with an arbitrary differentiable cost function of the trajectory, how is this a function of the noise parameter $\sigma$?

The key insight is to exploit the duality between any intermediate noisy trajectory $\bm{S}$ and the denoised trajectory at that noise level $D(\bm{S};\bm{C},\sigma)$. While $\bm{S}$ is clearly off the data manifold and not a physical trajectory, $D(\bm{S};\bm{C},\sigma)$ usually closely resembles a physical trajectory that is on the data manifold since it is trained to regress for the ground truth (Eqn. \ref{eqn:denoiser_loss}), even at a high $\sigma$ value. The denoised event and the noisy event converge at the limit $\sigma\rightarrow 0$. In this light, we approximate the constraint gradient score as:
\begin{align}
    \nabla_{\bm{S}}\log q(\bm{S};\bm{C},\sigma)\approx \lambda \frac{\partial}{\partial \bm{S}}\mathcal{L}\Big(D(\bm{S};\bm{C},\sigma)\Big)
    \label{eqn:def_constraint_score}
\end{align}
where $\mathcal{L}:\mathbb{R}^{N_a\times N_t\times N_f}\mapsto \mathbb{R}$ is an arbitrary cost function for the set of sampled trajectories, and $\lambda$ is a hyperparameter controlling the weight of this constraint.

In this work, we introduce two simple cost functions for trajectory controls: an attractor and a repeller. Attractors encourage the predicted trajectory at certain timesteps to arrive at certain locations. Repellers discourage interacting agents from getting too close to each other and mitigates collisions. We define the costs as:

\paragraph{Attractor cost}
\begin{align}
    \mathcal{L}_{\text{attract}}(D(\bm{S};\bm{C},\sigma))=\frac{\sum |(D(\bm{S};\bm{C},\sigma)-\bm{S}_{\text{target}})\odot \bm{M}_{\text{target}}|}{\sum|\bm{M}_{\text{target}}|+eps}
\end{align}
Where $\bm{S}_{\text{target}}\in\mathbb{R}^{N_a\times N_t\times N_f}$ are the target location tensor, and $\bm{M}_{\text{target}}$ is a binary mask tensor indicating which locations in $\bm{S}_{\text{target}}$ to enforce. $\odot$ denotes the elementwise product and $eps$ denotes an infinitesimal value to prevent underflow.

\paragraph{Repeller cost}

\begin{align}
    &\bm{A} = \texttt{max}\Big(\big(1-\frac{1}{r}\Delta(D(\bm{S};\bm{C},\sigma))\big)\odot(1-I), 0\Big) \\
    &\mathcal{L}_{\text{repell}}(D(\bm{S})) = \frac{\sum \bm{A}}{\sum (\bm{A} > 0) + eps}
\end{align}
Where $A$ is the per time step repeller cost. we denote the pairwise $L_2$ distance function between all pairs of denoised agents at all time steps as $\Delta(D(\bm{S};\bm{C},\sigma))\in\mathbb{R}^{N_a\times N_a\times N_t}$, identity tensor broadcast to all $N_t$ time steps $\bm{I}\in\mathbb{R}^{N_a\times N_a\times N_t}$, and repeller radius as $r$.

\paragraph{Constraint score thresholding}
To further increase the stability of the constrained sampling process, we propose a simple and effective strategy: constraint score thresholding (ST). From Eqn. \ref{eqn:score_vs_denoiser}, we make the observation that:
\begin{align}
    \sigma\nabla_{\bm{x}}\log p(\bm{x};\sigma)=(D(\bm{x},\sigma)-\bm{x})/\sigma=\bm{\epsilon},~\bm{\epsilon}\sim\mathcal{N}(\bm{0},\bm{I})
\end{align}
Therefore, we adjust the constraint score in Eqn. \ref{eqn:def_constraint_score} via an elementwise clipping function:
\begin{align}
    \nabla_{\bm{S}}\log q(\bm{S};\bm{C},\sigma):=\texttt{clip}(\sigma \nabla_{\bm{S}}\log q(\bm{S};\bm{C},\sigma), \pm 1)/\sigma
\end{align}
We ablate this design choice in Table \ref{tab:constraint}.
\subsection{Trajectory Representation Enhancements}\label{ssec:representation_enhancements}
\paragraph{Sample clustering}
While \ours{} learns an entire distribution of possible joint future trajectories from which we can draw an arbitrary number of samples, it is often necessary to extract a more limited number of representative modes from the output distribution. The Interaction Prediction challenge in Waymo Open Motion Dataset, for instance, computes metrics based on a set of 6 predicted joint futures across modeled agents. Thus, we need to generate a representative set from the larger set of sampled trajectories. 

To this end, we follow the trajectory aggregation method defined in  \cite{varadarajan2021multipath++} which performs iterative greedy clustering to maximize the probability of trajectory samples falling within a fixed distance threshold to an output cluster. We refer readers to \cite{varadarajan2021multipath++}  for details on the clustering algorithm.

In the joint agent prediction setting, we modify the clustering algorithm such that for each joint prediction sample, we maximize the probability that all agent predictions fall within a distance threshold to an output cluster.



\paragraph{PCA latent diffusion} Inspired by the recent success of latent diffusion models \cite{rombach2022high} for image generation, we utilize a compressed representation for trajectories using Principal Component Analysis (PCA). PCA is particularly suitable for representing trajectories, as trajectories are temporally and geometrically smooth in nature, and the trajectories can be represented by a very small set of components. Our analysis shows that a mere 3 components (for trajectories with $80\times2$ degrees of freedom) accounts for $99.7\%$ of all explained variance, though we use 10 components for a more accurate reconstruction. PCA representation has multiple benefits, including faster inference, better success with controlled trajectory, and perhaps most importantly, better accuracy and performance (see ablation studies in Sec. \ref{sec:ablation}).

First, as many ground truth trajectories include missing time steps (due to occlusion / agent leaving the scene), we use linear interpolation / extrapolation to fill in the missing steps in each trajectory. We uniformly sample a large population of $N_s=10^5$ agent trajectories, where each trajectory $\bm{s}_i\in\mathbb{R}^{N_t N_f}, i\in[1, N_s]$ is first centered around the agent's current location, rotated such that the agent's heading is in $+y$ direction, and flattened into a single vector. Denote this random subset of agent trajectories as $\bm{S}'\in\mathbb{R}^{N_s\times N_t N_f}$. We compute its corresponding principle component matrix (with whitening) as $W_{\text{pca}}\in\mathbb{R}^{N_p\times(N_t N_f)}$ where $N_p$ is the number of principle components to use, and its mean as $\bar{\bm{s}}'\in\mathbb{R}^{N_t N_f}$. We obtain the PCA and inverse PCA transformation for each trajectory $\bm{S}_i$ as:
\begin{align}
\hat{\bm{s}}_i = (\bm{s}_i-\bar{\bm{s}})W_{\text{pca}}^T \Leftrightarrow \bm{s}_i = \hat{\bm{s}}_i (W_{\text{pca}}^T)^{-1}+\bar{\bm{s}}
\end{align}
With the new representation, we have agent trajectories in Eqn. \ref{eqn:set_prob} in PCA space as $\bm{S}\in\mathbb{R}^{N_a\times N_p}$.

\section{Experiment and Results}


\subsection{PCA Mode Analysis}\label{ssec:pca_modal_analysis}

\begin{figure}
     \centering
     \begin{subfigure}[b]{0.48\linewidth}
         \centering
         \includegraphics[width=\linewidth,height=\linewidth]{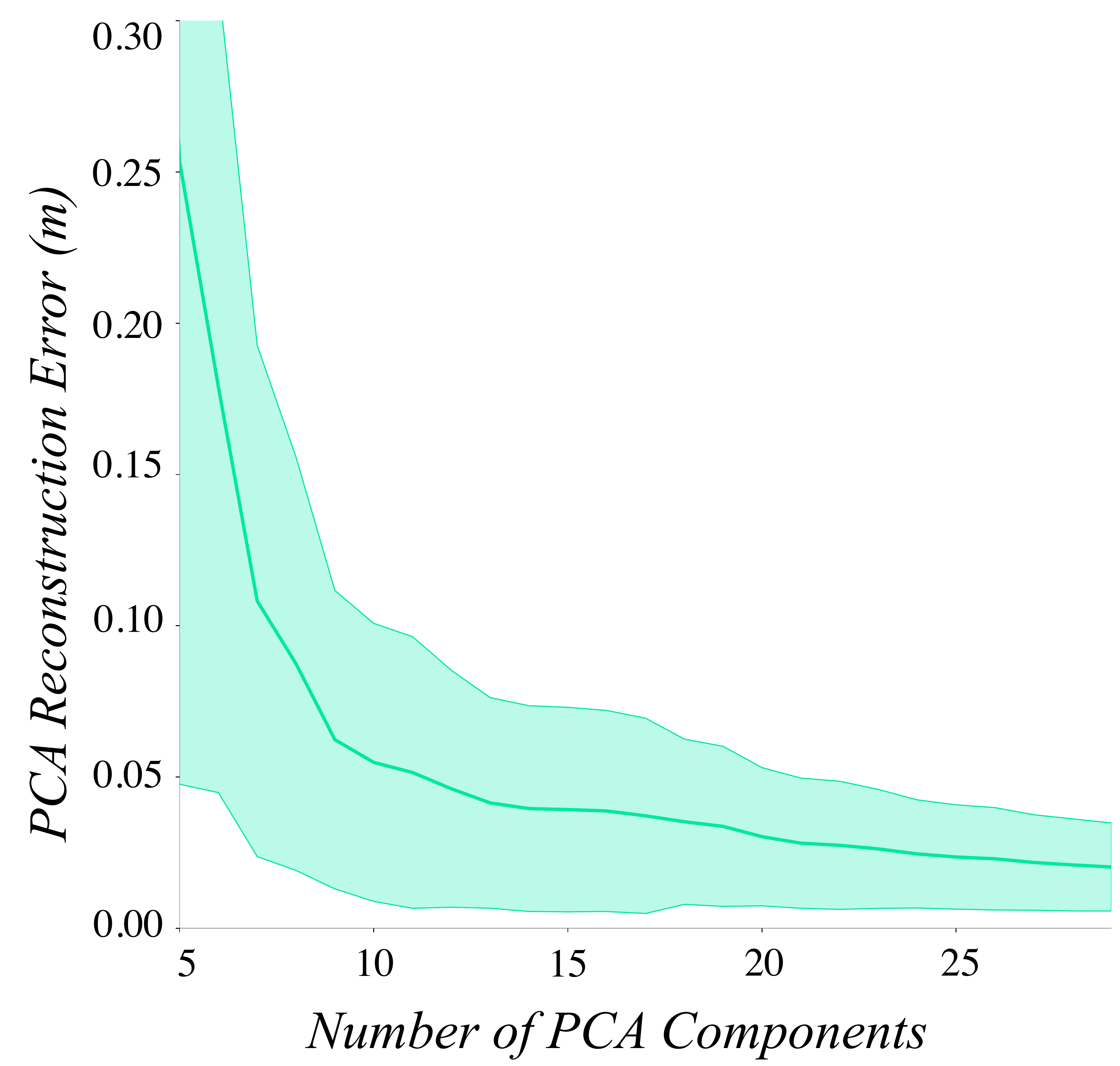}
         \subcaption{PCA trajectory reconstruction error vs number of PCA components.}
         \label{fig:num_pca_vs_error}
     \end{subfigure}
     \hfill
     \begin{subfigure}[b]{0.48\linewidth}
         \centering
         \includegraphics[width=\linewidth,height=\linewidth]{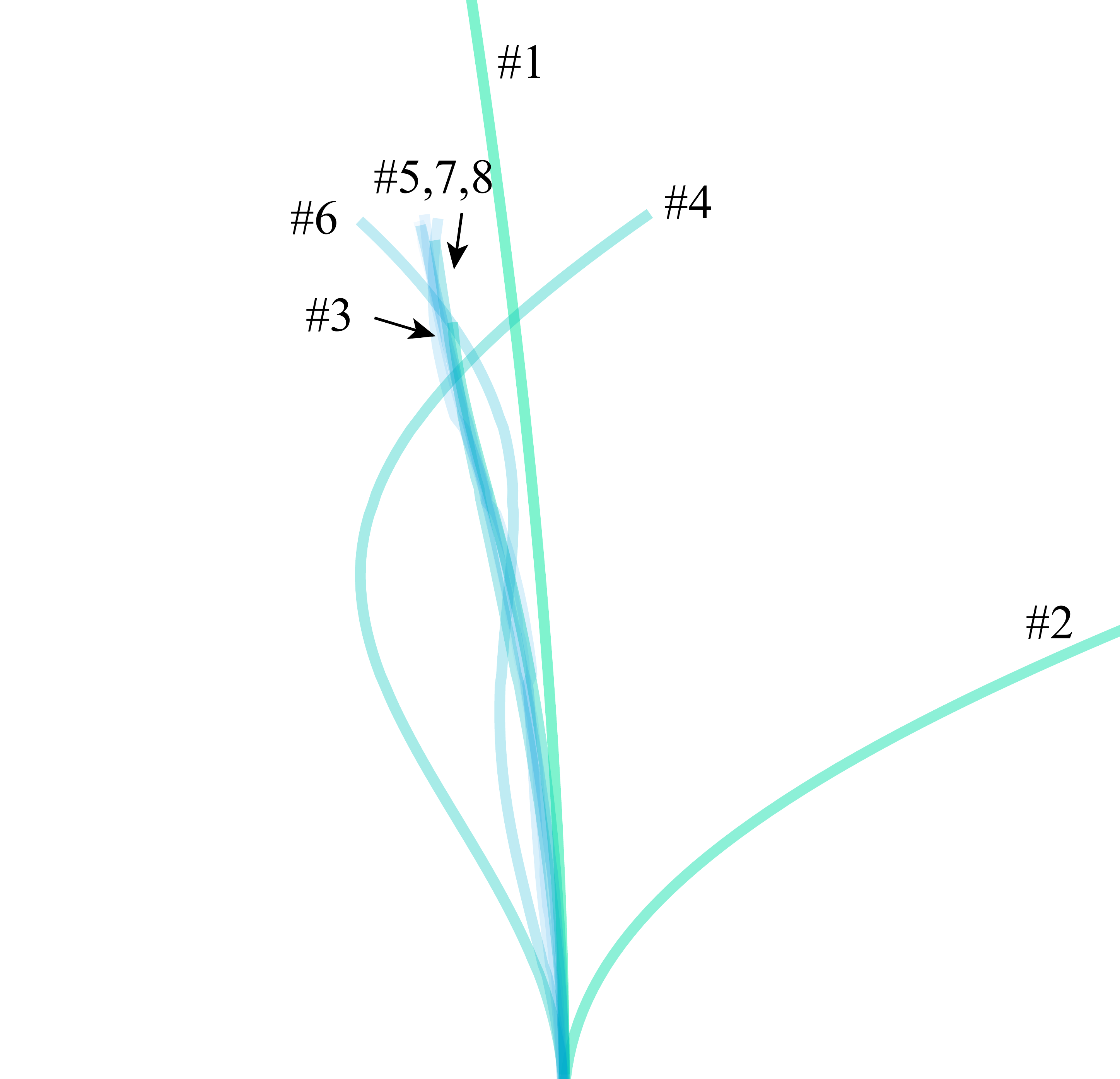}
         \caption{Visualization of the top-$10$ PCA components for trajectories.}
         \label{fig:viz_pca_components}
     \end{subfigure}
     \caption{Analysis of PCA representation for agent trajectories. (a) shows the average reconstruction error for varying numbers of principal components. (b) shows a visualization of the top-$10$ principal components. The higher modes representing higher frequencies are increasingly similar and have a small impact on the final trajectory.}
     \label{fig:pca_modal_analysis}
\end{figure}

To motivate our use of PCA as a simple and accurate compressed trajectory representation, we analyze the principal components computed from $N_s=10^5$ randomly selected trajectories from the Waymo Open Dataset training split. Fig. \ref{fig:num_pca_vs_error} shows the average reconstruction error per waypoint using increasing numbers of principal components. When keeping only the first 10 principal components, the average reconstruction error is 0.06 meters, which is significantly lower than the average prediction error achieved by state-of-the-art methods. This motivates PCA as an effective compression strategy, without the need for more complex strategies like autoencoders in \cite{rombach2022high}. 

We visualize the top-$10$ principal components in Fig. \ref{fig:viz_pca_components}. The higher order principal components are increasingly similar, and deviate only slightly from the dataset mean. These components represent high frequency trajectory information that are irrelevant for modeling, and may also be a result of perception noise.

\begin{figure*}[h!]
    \centering
    \includegraphics[width=\linewidth]{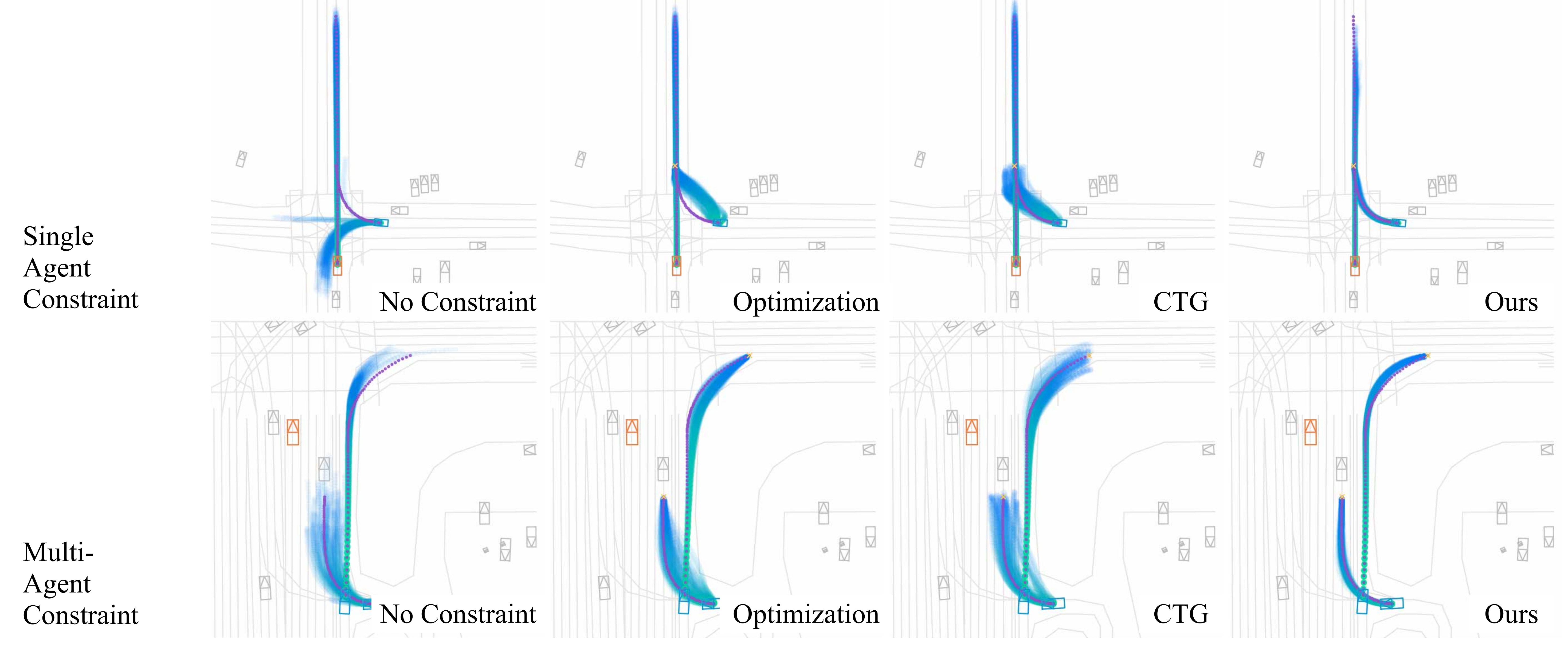}
    \caption{Qualitative results for controllable trajectory synthesis. We apply an attractor-based constraint (marked as \textcolor{amber500}{$\times$}) on the last point of the trajectory. Without any constraint at inference time, the initial prediction distributions from \ours{} (``No Constraint'') are plausible yet dispersed. While test time optimization of the predicted trajectories is effective at enforcing the constraints on model outputs, it deviates significantly from the data manifold, resulting in unrealistic outputs. Our method produces realistic and well-constrained results.}
    \label{fig:control_traj}
    \vspace{-1em}
\end{figure*}

\subsection{Multi-Agent Motion Prediction}
\label{ssec:womd_inter}
To evaluate \ours{}'s performance in the multi-agent prediction setting, we assess our method on the Waymo Open Dataset Interactive split, which contains pairs of agents in highly interactive and diverse scenarios \cite{ettinger2021large}. 

In Table \ref{table:womd_inter}, we report the main metrics for the Interactive split, as defined in \cite{ettinger2021large}. minSADE measures the displacement between the ground-truth future agent trajectories and the closest joint prediction (out of 6 joint predictions), averaged over the future time horizon and over the pair of interacting agents. minSFDE measures the minimum joint displacement error at the time horizon endpoint. SMissRate measures the recall of the joint predictions, with distance thresholds defined as a function of agent speed and future timestep. Finally mAP measures the joint Mean Average Precision based on agent action types, such as left-turn and u-turn. The reported metrics are averaged over future time horizons (3s, 5s, and 8s) and over agent types (vehicles, pedestrians, and cyclists) . 

Additionally, we report results for the Overlap metric \cite{wenjie2022} by measuring the overlap rate on the most likely joint prediction, which captures the consistency of model predictions, as consistent joint predictions should not collide.

Our model achieves state-of-the-art results, as shown in Table \ref{table:womd_inter}. While \ours{} and  Wayformer \cite{nayakanti2022wayformer} use the same backbone, our method performs significantly better across all metrics due to the strength of the diffusion head. Compared to JFP \cite{wenjie2022} on the test split, we demonstrate an improvement with respect to the minSADE and minSFDE metrics. For mAP, and Overlap, our method performs slightly worse than JFP, but outperforms all other methods.


\begin{table}[t]
\scriptsize
\setlength{\tabcolsep}{1mm}
\centering
\begin{tabular}{l|lccccccccccccc}
\specialrule{.2em}{.1em}{.1em}
& Method & Overlap & \scalebox{0.88}[1.0]{minSADE} & \scalebox{0.88}[1.0]{minSFDE} & \scalebox{0.88}[1.0]{sMissRate} & mAP \\
&  & ($\downarrow$) & ($\downarrow$) & ($\downarrow$) & ($\downarrow$) & ($\uparrow$) \\ \hline
\multirow{8}{1.5em}{Test} 
& LSTM baseline~\cite{ettinger2021large}    & - & 1.91 &	5.03 &	0.78 &	0.05 \\
& HeatIRm4~\cite{mo2021multi}               & - & 1.42 &	3.26 &	0.72 &	0.08 \\
& SceneTransformer(J)~\cite{ngiam21scene_transformer}   & - & 0.98 &	2.19 &	0.49 &	0.12 \\
& M2I~\cite{sun2022m2i}                     & - & 1.35 &	2.83 &	0.55 &	0.12 \\
& DenseTNT~\cite{gu21dense_tnt}             & - & 1.14 &	2.49 &	0.54 &	0.16 \\

& MultiPath++\cite{varadarajan2021multipath++} & - & 1.00 & 	2.33 & 	0.54 & 0.17 \\
& JFP~\cite{wenjie2022}                      & - & 0.88 & 1.99 & \bf{0.42} & \bf{0.21} \\
& MotionDiffuser (Ours)                     & - & \bf{0.86} & \bf{1.95} & 0.43 & 0.20 \\

\hline

\multirow{6}{1.5em}{Val}
& SceneTransformer(M)~\cite{ngiam21scene_transformer} & 0.091 & 1.12 & 2.60 & 0.54 & 0.09 \\
& SceneTransformer(J)~\cite{ngiam21scene_transformer} & 0.046 & 0.97 & 2.17 & 0.49 & 0.12 \\
& MultiPath++~\cite{varadarajan2021multipath++} & 0.064 & 1.00 & 2.33 & 0.54 & 0.18 \\
& JFP~\cite{wenjie2022} & \bf{0.030} & 0.87 & 1.96 & \bf{0.42} & \bf{0.20} \\
& Wayformer~\cite{nayakanti2022wayformer} & 0.061 & 0.99 & 2.30 & 0.47 & 0.16 \\
& MotionDiffuser (Ours)                     & 0.036 & \bf{0.86} & \bf{1.92} & \bf{0.42} & 0.19 \\

\specialrule{.1em}{.05em}{.05em}

\end{tabular}
\caption{WOMD Interactive Split: we report scene-level joint metrics numbers averaged for all object types over t = 3, 5, 8 seconds. Metrics minSADE, minSFDE, SMissRate, and mAP are from the benchmark \cite{ettinger2021large}. Overlap is defined in \cite{wenjie2022}.}
\label{table:womd_inter}
\end{table}

\vspace{-.5em}
\subsection{Controllable Trajectory Synthesis}
\label{ssec:controllable_traj_synthesis}

\begin{table}[t]
\scriptsize
\setlength{\tabcolsep}{1mm}
\begin{tabular}{@{}l|rrr|rrrr@{}}
\toprule
                        & \multicolumn{3}{c|}{Realism ($\downarrow$)}               & \multicolumn{4}{c}{Constraint Effectiveness} \\ \cmidrule(l){2-8} 
\multirow{-2}{*}{Method} &
  \multicolumn{1}{l}{\scalebox{0.7}[1.0]{minSADE}} &
  \multicolumn{1}{l}{\scalebox{0.7}[1.0]{meanSADE}} &
  \multicolumn{1}{l|}{\scalebox{0.7}[1.0]{Overlap}} &
  \multicolumn{1}{l}{\scalebox{0.7}[1.0]{minSFDE ($\downarrow$)}} &
  \multicolumn{1}{l}{\scalebox{0.7}[1.0]{meanSFDE ($\downarrow$)}} &
  \multicolumn{1}{l}{\scalebox{0.8}[1.0]{SR2m ($\uparrow$)}} &
  \multicolumn{1}{l}{\scalebox{0.8}[1.0]{SR5m ($\uparrow$)}} \\ \midrule
\scalebox{0.8}[1.0]{No Constraint}           & 1.261 & 3.239 & 0.059 & 2.609      & 8.731      & 0.059     & 0.316     \\ \midrule
\multicolumn{8}{l}{\textit{Attractor} (to GT final point)}  \\
\scalebox{0.8}[1.0]{Optimization}            & 4.563 & 5.385 & 0.054 & \textit{0.010}      &\textbf{ 0.074}      & \textbf{1.000}         & \textbf{1.000}         \\
GTC\cite{zhong2022guided} & 1.18  & \textbf{1.947} & 0.057 & 0.515      & 0.838      & 0.921     & 0.957     \\
Ours (-ST) & \textit{1.094} & \textit{2.083} & 0.042 & 0.627      & 1.078      & 0.913     & 0.949     \\
Ours                    & \textbf{0.533} & 2.194 & \textit{0.040}  & \textbf{0.007}      & \textit{0.747}      & \textit{0.952}     & \textit{0.994}     \\ \midrule
\multicolumn{8}{l}{\textit{Repeller} (between the pair of agents)}  \\
Ours & 1.359 & 3.229 & \textbf{0.008}  & 2.875 & 8.888 & 0.063 & 0.317 \\
\bottomrule
\end{tabular}
\caption{Quantitative validation for controllable trajectory synthesis. We enforce the attractor or repeller constraints in Sec. \ref{ssec:constrained_traj_sample}.}
\label{tab:constraint}
\vspace{-1em}
\end{table}
We experimentally validate the effectiveness of our controllable trajectory synthesis approach. In particular, we validate the attractor and repeller designs proposed in Sec. \ref{ssec:constrained_traj_sample}. We continue these experiments using the Interactive Split from Waymo Open Motion Dataset. In experiments for both the attractor and the repeller, we use the same baseline diffusion model trained in Sec. \ref{ssec:womd_inter}. We randomly sample 64 trajectories from the predicted distribution. We report our results in Table \ref{tab:constraint}. We measure min/mean ADE/FDE and overlap metrics, following Sec. \ref{ssec:womd_inter}. The mean metrics computes the mean quantity over the 64 predictions.

For the attractor experiment, we constrain the last point of all predicted trajectories to be close to the last point in the ground truth data. Therefore, min/meanSADE serves as a proxy for the realism of the predictions and how closely they stay to the data manifold. For baselines, we compare to two approaches: ``Optimization" directly samples the trajectories from our diffusion model, followed by a post processing step via Adam optimizer to enforce the constraints. ``CTG" is a reimplementation of the sampling method in a concurrent work \cite{zhong2022guided} that performs an inner optimization loop to enforce constraints on the denoised samples during every step of the diffusion process. See Table \ref{tab:constraint} for detailed results. Although trajectory optimization after the sampling process has the strongest effect in enforcing constraints, it results in unrealistic trajectories. With our method we have a high level of effectiveness in enforcing the trajectories, second only to optimization methods, while maintaining a high degree of realism. Additionally we show qualitative comparisons for the optimized trajectories in Fig. \ref{fig:control_traj}.

For the repeller experiment, we add the repeller constraint (radius $5m$) between all pairs of modeled agents. We were able to significantly decrease overlap between joint predictions by an order of magnitude, demonstrating its effectiveness in repelling between the modeled agents.

\vspace{-.5em}
\section{Ablation Studies}
\label{sec:ablation}

\begin{table}[t]
\small
\setlength{\tabcolsep}{1mm}
\centering
\begin{tabular}{lccccccccccccc}
\specialrule{.2em}{.1em}{.1em}
Method  & \scalebox{0.9}[1.0]{minSADE($\downarrow$)} & \scalebox{0.9}[1.0]{minSFDE($\downarrow$)} & \scalebox{0.9}[1.0]{SMissRate($\downarrow$)} \\ \hline
Ours (-PCA)             & 1.03 & 2.29 & 0.53 \\
Ours (-Transformer)  & 0.93 & 2.08 & 0.47 \\
Ours (-SelfAttention)  & 0.91 & 2.07 & 0.46 \\
MotionDiffuser (Ours)   & 0.88 & 1.97 & 0.43 \\

\specialrule{.1em}{.05em}{.05em}

\end{tabular}
\caption{Ablations on WOMD Interactive Validation Split. We ablate components of the denoiser architecture, and the PCA compressed trajectory representation.}
\label{table:ablation}
\vspace{-2em}
\end{table}

We validate the effectiveness of our proposed Score Thresholding (ST) approach in Table \ref{tab:constraint}, with \textit{Ours(-ST)} denoting the removal of this technique, resulting in significantly worse constraint satisfaction.

Furthermore, we ablate critical components of the \ours{} architecture in Table \ref{table:ablation}. We find that using the uncompressed trajectory representation \textit{Ours(-PCA)} degrades performance significantly. Additionally, replacing the Transformer architecture with a simple MLP \textit{Ours(-Transformer)} reduces performance. We also ablate the self-attention layers in the denoiser architecture \textit{Ours(-SelfAttention)}, while keeping the cross-attention layers (to allow for conditioning on the scene context and noise level). This result shows that attention between modeled agents' noisy future trajectories is important for generating consistent joint predictions. Note that \ours{}'s performance in Table \ref{table:ablation} is slightly worse than Table \ref{table:womd_inter} due to a reduced Wayformer encoder backbone size.

\vspace{-.5em}
\section{Conclusion and Discussions}
\label{sec:conclusion}
In this work, we introduced \ours{}, a novel diffusion model based multi-agent motion prediction framework that allows us to learn a diverse, multimodal joint future distribution for multiple agents. We propose a novel transformer-based set denoiser architecture that is permutation invariant across agents. Furthermore we propose a general and flexible constrained sampling framework, and demonstrate the effectiveness of two simple and useful constraints - attractor and repeller. We demonstrate state-of-the-art multi-agent motion prediction results, and the effectiveness of our approach on Waymo Open Motion Dataset.

Future work includes applying the diffusion-based generative modeling technique to other topics of interest in autonomous vehicles, such as planning and scene generation.

\paragraph{Acknowledgements} We thank Wenjie Luo for helping with the overlap metrics code, Ari Seff for helping with multi-agent NMS, Rami Al-RFou, Charles Qi and Carlton Downey for helpful discussions, Joao Messias
for reviewing the manuscript, and anonymous reviewers.

{\small
\bibliographystyle{ieee_fullname}
\bibliography{Reference}
}

\end{document}


\title{\ours{}: Controllable Multi-Agent Motion Prediction using Diffusion}

\author{Chiyu ``Max" Jiang$^{*}$ \qquad Andre Cornman$^{*}$ \qquad Cheolho Park \\
\qquad Ben Sapp \qquad Yin Zhou \qquad Dragomir Anguelov \\
{\tt\small {$*$ equal contribution}} \\
Waymo LLC \\
}
\maketitle

\setlength{\abovedisplayskip}{0.2\abovedisplayskip}
\setlength{\belowdisplayskip}{0.2\belowdisplayskip}
\setlength{\abovecaptionskip}{0.2\abovecaptionskip}
\setlength{\belowcaptionskip}{0.2\belowcaptionskip}

\section{Additional Visualizations}

\begin{figure}[H]
    \centering
    \includegraphics[width=0.85\linewidth]{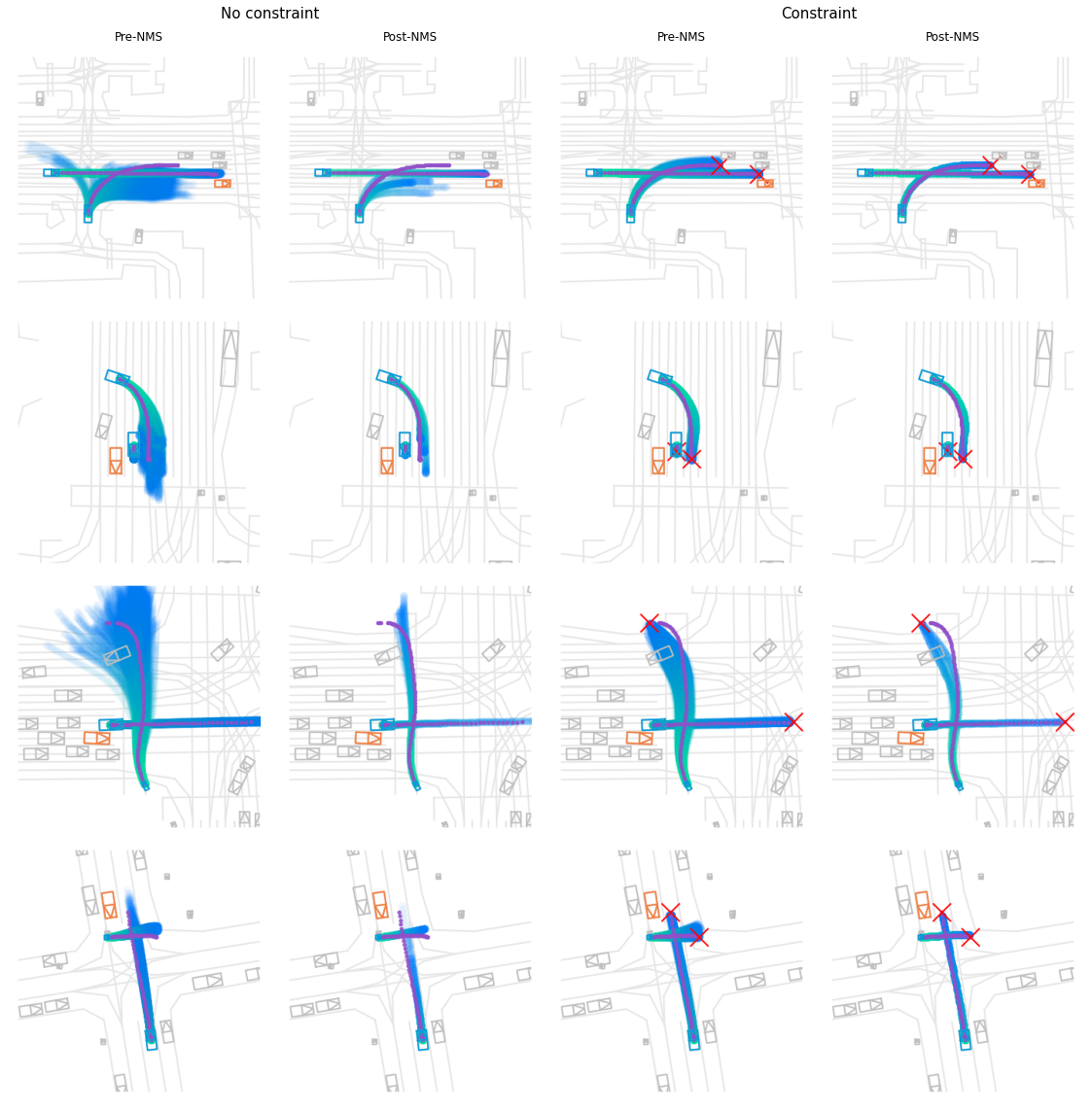}
\end{figure}

\begin{figure}[H]
    \centering
    \includegraphics[width=0.85\linewidth]{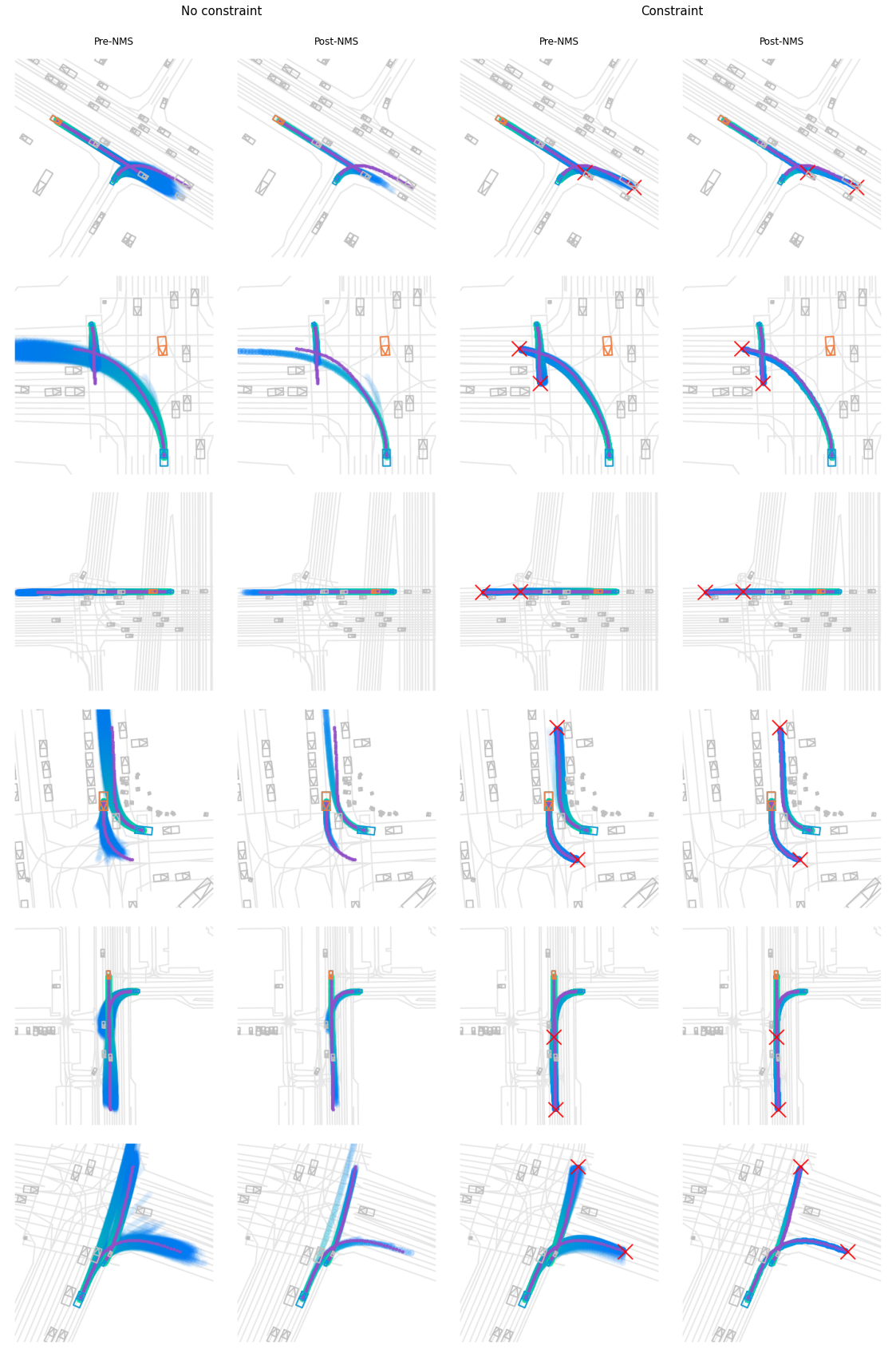}
\end{figure}

\begin{figure}[H]
    \centering
    \includegraphics[width=0.85\linewidth]{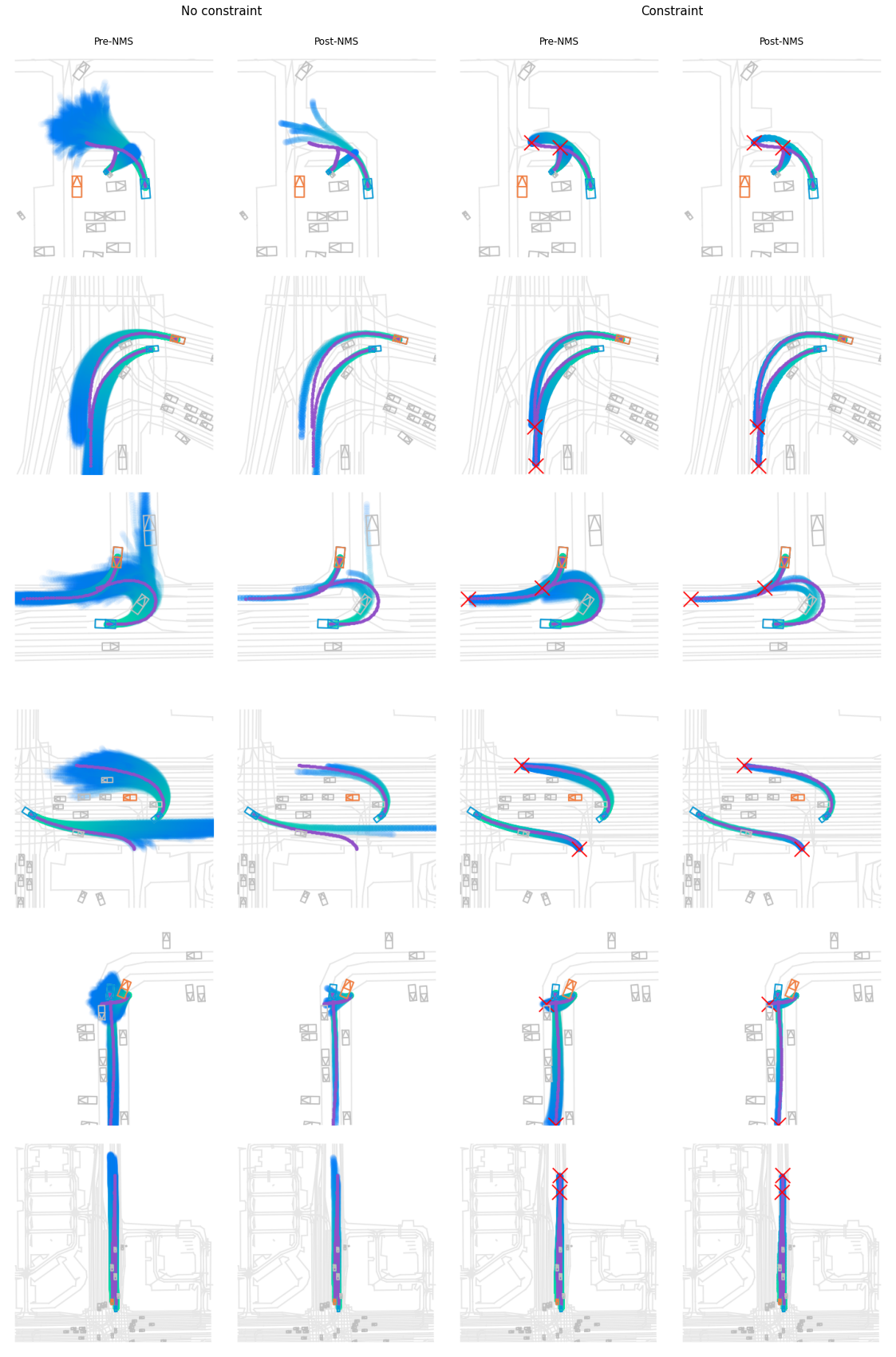}
\end{figure}

\section{Implementation Details}

\ours{} is trained on the Waymo Open Motion Dataset using 32 TPU shards for $2*10^6$ training steps. We use the ADAMW optimizer \cite{Loshchilov2017adamw} with weight decay coefficient of 0.03. The learning rate is set to $5 * 10^{-4}$, with $10^4$ warmup steps and linear learning rate decay. \ours{} uses the Wayformer \cite{nayakanti2022wayformer} encoder backbone, with 128 latent embeddings, each with hidden size of 256. Because the Wayformer encoder is agent centric, we append each agent’s position and heading (relative to the ego vehicle) to its corresponding context vectors.

Our transformer denoiser architecture uses 4 layers of self-attention and cross-attention blocks. Each attention layer has a hidden size of 256 and an intermediate size of 1024. ReLU activation is used in all transformer layers. We embed the noise level using 128 random fourier features.

We can flexibly denoise $N$ random noise vectors during training and inference. We use $N = 128$ during training and $N = 256$ during inference (before applying clustering).

\section{Network Preconditioning}

We follow the network preconditioning framework from \cite{karras2022elucidating}, which defines the denoiser $D_{\bm{\theta}}$ as:
\begin{align}
    D_{\bm{\theta}}(\bm{x};\bm{c},\sigma)=\cskip(\sigma)\bm{x}+\cout(\sigma)F_{\bm{\theta}}(\cin(\sigma)\bm{x};\bm{c},\cnoise(\sigma))
\end{align}

$\cin(\sigma)$ scales the network input, such that the training inputs to $F_{\bm{\theta}}$ have unit variance.
\begin{align}
    \cin(\sigma) = 1 / \sqrt{\sigma^2 + \sigma^{2}_{data}}
\end{align}

$\cskip(\sigma)$ modulates the skip connection and is defined as:
\begin{align}
    \cskip(\sigma) = \sigma^{2}_{data} / (\sigma^2 + \sigma^{2}_{data})
\end{align}

$\cout(\sigma)$ modulates the network output and is defined as:
\begin{align}
    \cout(\sigma) = \sigma \cdot \sigma_{data} / \sqrt{\sigma^2 + \sigma^{2}_{data}}
\end{align}

Finally $\cnoise(\sigma)$ scales the noise level, and is defined as:
\begin{align}
    \cnoise(\sigma) = \frac{1}{4} \ln{\sigma}
\end{align}

For all our experiments, we set $\sigma_{data} = 0.5$.

\section{Inference Latency}
 We report our model's inference latency over a varying number of sampling steps $T$ in Table \ref{table:latency}. We use a single V100 GPU, with batch size of 1.
\begin{table}[H]
\small
\setlength{\tabcolsep}{1mm}
\centering
\begin{tabular}{lccccccccccccc}
\specialrule{.2em}{.1em}{.1em}

Method  & \scalebox{0.9}[1.0]{Latency (ms)} & \scalebox{0.9}[1.0]{minSADE($\downarrow$)} & \scalebox{0.9}[1.0]{minSFDE($\downarrow$)} & \scalebox{0.9}[1.0]{SMissRate($\downarrow$)} \\ \hline
Ours ($T=8$)   & 101.0 & 0.91 & 2.06 & 0.47 \\ 
Ours ($T=16$)  & 203.7 & 0.88 & 1.96 & 0.44 \\
Ours ($T=32$)  & 408.5 & 0.88 & 1.97 & 0.43 \\

\specialrule{.1em}{.05em}{.05em}

\end{tabular}
\caption{Model inference latency vs. quality for WOMD Interactive Validation Split.}
\label{table:latency}
\end{table}

{\small
\bibliographystyle{ieee_fullname}

\bibliography{Reference}
}